\documentstyle{llncs}

\def\argmax{{\rm argmax}}

\def\argmin[{{\rm argmin}}

\def\x{{\bf x}}
\def\y{{\bf y}}
\def\s{{\bf s}}
\def\p{{\bf p}}

\def\z{{\bf z}}

\def\b{{\bf b}}
\def\a{{\bf a}}
\def\e{{\bf e}}
\def\X{{\bf X}}

\def\c{{\bf c}}

\def\0{{\bf 0}}
\def\f{{\bf f}}
\def\g{{\bf g}}

\def\P{{\bf P}}
\def\Q{{\bf Q}}

\begin{document}
\title{Log-Optimal Portfolio Selection Using the
Blackwell Approachability Theorem}

\author{Vladimir V'yugin}

\institute{Institute for Information Transmission Problems,
Russian Academy of Sciences,
Bolshoi Karetnyi per. 19, Moscow GSP-4, 127994, Russia\\
email: vyugin@iitp.ru}

\maketitle

\begin{abstract}
The goal of this paper is to apply game-theoretic methods
to the classical problem of an optimal portfolio construction.
We present a method for constructing a log-optimal portfolio
using the well-calibrated forecasts of return vectors.
Dawid's notion of calibration and the Blackwell approachability theorem
are used for computing the well-calibrated forecasts.
We select a portfolio using this ``artificial'' probability distribution
of return vectors.
Our portfolio performs asymptotically at least as well as any
stationary portfolio that redistributes the investment at each round using
a continuous function of side information.
Unlike in classical mathematical finance theory, no stochastic assumptions
are made about return vectors.
\end{abstract}

\section{Introduction}\label{intr-1}

The model of stock market considered in this paper is the one studied,
among others, by Breiman~\cite{Bre61}, Algoet and Cover~\cite{CovAlC88},
and Cover~\cite{Cov91}; see also Gy\"orpfy et al.~\cite{Gyo2006}.

Consider an investor who can access $k$ financial instruments (asset,
bond, cash, return of a game, etc.), and who can rebalance his wealth in each
round according to a portfolio vector ${\b} = (b(1),\dots, b(k))$.
The evolution of the market in time is represented by a sequence of return
vectors (market values) $\x_1,\x_2,\dots$.  A return vector $\x = (x(1),\dots ,x(k))$
is a vector of $k$ nonnegative numbers representing price relatives
for a given trading period. That is, the $j$th component $x(j)$
of $\x$ expresses the ratio of the closing and opening prices of asset $j$.
In other words, $x(j)$ is the factor by which capital invested in the $j$th
asset grows during the trading period. We suppose that these components
are bounded $x(j)\in [\lambda_1,\lambda_2]$
for all $1\le j\le k$, where $0<\lambda_1<\lambda_2<\infty$.

We assume that
the assets are arbitrarily divisible, and they are available for buying or
for selling in unbounded quantities at the current price at any given
trading period; there are no transaction costs,
The investor is allowed to diversify his capital at the beginning of each
trading period according to a portfolio vector $\b = (b(1),\dots ,b(k))$.
The $j$th component $b(j)$ of the vector $\b$ denotes the
proportion of the investor's capital invested in asset $j$.
We assume that $b(j)\ge 0$ for all $j$ and $\sum\limits_{j=1}^k b(j)=1$.
This means that the investment strategy is self-financing and consumption
of capital is excluded. The nonnegativity of the components of $\b$ means
that short selling and buying stocks on margin are not permitted.
Let $\Gamma$ denotes the simplex of portfolio vectors $\b$.

Let $S_0$ denote investor''s initial capital.
Then, at the end of the first trading period, investor''s wealth becomes
$$
S_1=S_0(\b\cdot\x)=S_0\sum\limits_{j=1}^k b(j)x(j),
$$
where $"\cdot"$ denotes the inner product.

Starting with an initial wealth $S_0$, after $T$ trading periods,
an ``investment strategy'' $\b_1,\dots ,\b_T$ achieves a wealth
$$
S_T=S_0\prod_{t=1}^T (\b_t\cdot\x_t).
$$
For simplicity, in what follows we assume that $S_0=1$.

In modeling the behavior of the evolution of the market, two main approaches
have been considered in the theory of sequential investment.
In the probabilistic approach, we assume that return vectors $\x_t$
are realization of a sequence of random process $\X_t$,
where $t=1,2,\dots$, and describe a statistical model.

If a market process $\X_t$ is memoryless, i.e., it is a sequence of
independent and identically distributed (i.i.d.) random return vectors
then it was shown by Morvai~\cite{Mor91} that the portfolio\footnote{In what follows
$\log$ denotes logarithm on the base 2}
$
\b^*=\argmax_{\b}E(\log(\b\cdot\X_1))
$
is asymptotically optimal in the following sense: for any portfolio vector $\b$
with finite
\footnote
{
That is true when $X_i$ are uniformly bounded.
}
$E((\log(\b\cdot\X_1))^2)$,
a condition of optimality
$
\liminf\limits_{T\to\infty}\frac{1}{T}\log\frac{S_T^*}{S_T}\ge 0
$
holds almost surely, where $S_T^*=\prod_{t=1}^T (\b^*\cdot\X_t)$ and
$S_T=\prod_{t=1}^T (\b\cdot\X_t)$.

But i.i.d. model is insufficient if the return
vectors of different trading periods have a statistical dependence,
which seems to be the case in real-world markets. In general case, we
consider an arbitrary random process $\X_1,\X_2,\dots$ generating return
vectors $\x_1,\x_2,\dots$.

Algoet~\cite{Alg92} and Cover~\cite{CovAlC88} have constructed
so-called log-optimum portfolio strategy.
Let $\X_1,\X_2,\dots$ be an arbitrary stationary and ergodic process.
Denote $\X_1^{t-1}=\X_1,\dots,\X_{t-1}$. A function $\b^*(\cdot)$ is said to be
a log-optimal portfolio strategy if
\begin{eqnarray}
E(\log(\b^*(\X_1^{t-1})\cdot\X_t))=\max\limits_{\b(\cdot)}
E(\log(\b(\X_1^{t-1})\cdot \X_t))
\label{log-optimal-1}
\end{eqnarray}
for all $t$.

Let $S^*_T$ denote a wealth achieved by a log-optimum portfolio strategy
$\b^*(\cdot)$ after $T$ trading periods. Then for any stationary and ergodic
process $\X_t$ and for any other investment strategy $\b(\cdot)$,
\begin{eqnarray}
\liminf\limits_{T\to\infty}\frac{1}{T}\log\frac{S^*_T}{S_T}\ge 0
\label{asympt-1}
\end{eqnarray}
almost surely. Such a strategy is also called universal with respect to the process
$\X_1,\X_2,\dots$.

Moreover, Algoet~\cite{Alg92} and Gy\"orpfy and Sch\"afer~\cite{GoS2003}
have shown that
there exists a strategy uniformly universal with respect to the class of all
stationary and ergodic processes. This means that
a strategy $\b^*(\cdot)$ exists such that for any stationary and
ergodic process $\X_1,\X_2,\dots$ asymptotic inequality (\ref{asympt-1})
holds almost surely.
Gy\"orfi and Sch\"afer called this scheme a histogram-based investment
strategy. Gy\"orpfy~\cite{Gyo2006} extended this result to the
kernel-based case. 
The rate of convergence has not been studied in these papers.

Another, ``worst-case'', approach allows a return sequence
$\x_1,\x_2,\dots$ to take completely arbitrary values, and no
stochastic model is imposed on the mechanism generating the
price relatives. This approach was pioneered by
Cover~\cite{Cov91}. Cover has shown that there exists an
investment strategy $\b^*_t$ (so-called universal portfolio)
that perform almost as well as the best portfolio in the sense
that for any return sequence $\x_1, \x_2,\dots$, $ S^*_T\ge
cT^{-\frac{k-1}{2}}S_T(\b) $ for all $T$, where $c$ is a
positive constant (depending on $k$),
$S^*_T=\prod_{t=1}^T(\b^*_t\cdot\x_t)$ is the wealth achieved
by of the universal portfolio strategy, and
$S_T(\b)=\prod_{t=1}^T(\b\cdot\x_t)$ is the wealth achieved by
arbitrary constant portfolio $\b$. The universal portfolio is
defined as the mixture $\b^*_T=\int\b P_{T-1}(d\b)$, where
$P_{T-1}(d\b)=(\prod_{t=1}^{T-1}(\b\cdot\x_t)/Z)D_{1/2}(d\b)$,
$Z=\int\prod_{t=1}^{T-1}(\b\cdot\x_t) D_{1/2}(d\b)$, and
$D_{1/2}$ is the $1/2$-Dirichlet distribution on $\Gamma$.

Further development of this approach see
in Cover and Ordentlich~\cite{CoO96}, Vovk and Watkins~\cite{VoW98},
Blum and Kalai~\cite{BlC99}, and so on. In this approach
the achieved wealth is compared with that of the best in a class of reference
strategies. The class of reference strategies considered by Cover~\cite{Cov91}
is that the class of all constant portfolios defined by all vectors
$\b\in\Gamma$. Cover and Ordentlich~\cite{CoO96} extended this
method for a case where side information in form of states from a finite set
can be used by a reference strategy.

%Using prediction with expert advice approach,
%Vovk~\citet{Vov2006} constructed a prediction strategy
%that perform almost as well as an arbitrary prediction strategy presented by
%a stationary continuous function of the side information. This result can be
%applied for construction of the portfolio log-optimal with respect to the reference
%class consisting of all arbitrary continuous portfolios $\b(\z_t)$.

The advantage of this worst-case approach is that it avoids imposing
statistical models on the stock market and the results hold for all possible
sequences $\x_1, \x_2,\dots$. In this sense this approach is extremely robust.

In Section~\ref{main-1}, we follow the combined "worst-case" and
stochastic approach. We construct an ``artificial probability
distribution'' of return vectors.
This distribution is defined by the well-calibrated forecasts
in Dawid~\cite{Daw82},~\cite{Daw85} sense; these forecasts
are constructed using the game-theoretic Blackwell approachability
theorem. No stochastic assumptions are made about return vectors
for constructing such forecasts. We construct a log-optimal portfolio
by scheme (\ref{log-optimal-1}), where the mathematical expectation $E$
is over the probability distribution defined by the well-calibrated forecasts.
Our log-optimal portfolio performs asymptotically at least as well as any
stationary portfolio that redistribute the investment at each round using
a continuous function of a side information. In Section~\ref{rate-1}
we discuss a rate of convergence of performance condition.

This approach is not new in the game theory.
For example, Foster and Vohra~\cite{FoV97},~\cite{FoV98}
have presented a ``calibrated'' forecasting scheme which is consistent
in hindsight. It allows the agent to choose
the best response to the predicted outcome. See also,
Mannor and Shimkin~\cite{MaSt2009}.

The goal of this paper is to apply these game-theoretic methods
to the classical problem of an optimal portfolio construction in order
to obtain a futher generalization of results presented in
the papers~\cite{GoS2003},~\cite{Gyo2006}.
Unlike most previous work we use a more broad reference class of
investor strategies -- we compare the performance of
our portfolio strategy with stationary investor''s strategies defined by
continuous
functions $\b_t=\b(\z_t)$ for all $t$, where $\z_t$ is a side information.

In Section~\ref{discr-1} we show that if $\b(\z_t)$ is allowed to be 
discontinuous, we cannot prove
asymptotic optimality of our portfolio strategy $\b_t^*$.

\section{Main result}\label{main-1}

{\bf Blackwell approachability theorem}.
We will define our randomized strategy for universal portfolio selection
using the Blackwell approachability theorem.

Recall some standard notions of the theory of games.
Consider a game between two players with finite sets of their moves
(pure strategies) $I=\{s_1,\dots ,s_N\}$ and $J=\{a_1,\dots ,a_M\}$.
A mixed strategy of the first player is a probability distribution on $I$
presented by a vector $\P=(p(1),\dots ,p(N))$, where $p(1)+\dots p(N)=1$ and
$p(i)\ge 0$
for all $i$. Denote by $P(I)$ a set of all mixed strategies of the first
player. Analogously, let $P(J)$ be the set of all mixed strategies of
the second player.

Consider an infinitely repeatable game, where at each round $t$ the first
player announces a mixed strategy $\P_t\in P(I)$ and the second player
announces a pure strategy $j_t\in J$. After that, the first player
pick up $i_t\in I$ distributed by $\P_t$ and receives a payoff $f(i_t,j_t)$.

The players can announce their moves independently or,
in the adversarial setting, where the second player announces
an element $j_t$ after the first player announces $\P_t$.
%In the adversarial setting, the second player can use $\P_t$ in his
%strategy for choosing $j_t$.

By a randomized online strategy of the first player we mean an infinite
sequence $\P_1,\P_2,\dots$ of his mixed strategies, where each $\P_t$
is a conditional (with respect to past moves $i_1,j_1,\dots ,i_{t-1},j_{t-1}$)
probability distribution on $I$.

Let a vector-valued payoff function $\f(s_i,a_j)\in {\cal R}^d$ be given,
where $d\ge 1$, $s_i\in I$, and $a_j\in J$. As usual,
$\f(\P,a_j)=\sum\limits_{i=1}^N\f(s_i,a_j)p(i)$ and
$\f(\P,\Q)=\sum\limits_{i=1}^N\sum\limits_{j=1}^M\f(s_i,a_j)p(i) q(j)$,
where $\P=(p(1),\dots ,p(N))$ and $\Q=(q(1),\dots ,q(M))$ are elements of
$P(I)$ and $P(J)$ respectively.

Note that the Blackwell theorem (see Theorem~\ref{approach-2} below) holds for $l_2$ norm
in ${\cal R}^d$. In some special case we also consider $l_1$
norm $\|\cdot\|_1$. In all other cases we use $l_2$ norm $\|\cdot\|$.
The choice of the norm, $l_1$, $l_2$ or $l_\infty$, is irrelevant at this stage,
since all norms are equivalent on finite-dimensional spaces. The difference
takes place in Section~\ref{rate-1}, where we compute a rate of convergence
in Theorem~\ref{opp-1}.

For any subset $U\subseteq {\cal R}^d$ and any vector
$\x\in {\cal R}^d$, the distance from $\x$ to $U$ is defined
$
{\rm dist}(\x,U)=\inf\limits_{\y\in U}\|\x-\y\|.
$

Following Blackwell a set $U\subseteq {\cal R}^d$ is called {\it approachable}
if a randomized online strategy $\P_1,\P_2,\dots$ of the first
player exists such that
\begin{eqnarray*}
\lim\limits_{T\to\infty}{\rm dist}\left(\frac{1}{T}
\sum\limits_{t=1}^T \f(i_t,j_t),U\right)=0
\end{eqnarray*}
holds for $P$-almost all sequences $i_1,i_2,\dots$ of the first player moves
regardless of how the second player chooses $j_1,j_2,\dots$,
where $P$ is an overall probability distribution generated by $\P_1,\P_2,\dots$.

Blackwell~\cite{Ble56} proposed a generalization of the
minimax theorem for the case of the vector-payoff functions. In particular,
he proved the following theorem.
\begin{theorem}\label{approach-2}
A closed convex subset $U\subseteq {\cal R}^d$ is approachable by the first
player if and only if for every mixed strategy $\Q\in P(J)$ of the second
player a mixed strategy $\P\in P(I)$ of the first player exists such that
$\f(\P,\Q)\in U$.
\end{theorem}
We apply this theorem for the log-optimal portfolio selection.

{\bf Optimal portfolio construction}.
We consider the market process in the game-theoretic framework
as a game between two players: {\it Market} and {\it Investor}.

In deterministic adversarial setting,
the market process is described as follows.
At the beginning of each time period $t$, {\it Investor}, observing
all past moves and a side information
$\z_t$ which is an element of some compact
metric space $C$, chooses a portfolio vector $\b_t$. At the end
of this period {\it Market}, observing all past moves, chooses
a return vector $\x_t$. After that,
{\it Investor} updates his wealth $S_t=S_{t-1}\cdot(\b_t\cdot\x_t)$.
For simplicity, we assume that $C$ is some closed interval
in $\cal R$.

A strategy of {\it Investor} is called stationary if
it is defined by a function from the set of all signals to the simplex of
all portfolios: $\b_t=\b(\z_t)$ for all $t$.

We discretize all basic sets used in the game. For any $K$, let
$\tilde C=\{\c_1,\dots ,~\c_K\}$ be a finite grid in $C$.
For every $\z\in C$ an $\c_j\in\tilde C$ exists such that
$\|\z-\c_j\|<\nu$, where
$\nu$ is a level of precision depending on $K$.

Recall that for any return vector $\x=(x(1),\dots ,x(k))$,
$x(i)\in [\lambda_1,\lambda_2]$,
where $\lambda_1$ and $\lambda_2$ are real numbers such that
$0<\lambda_1<\lambda_2$.

For any $M$, define a finite grid
$\tilde A=\{\a_1,\dots ,\a_M\}$ in the set $[\lambda_1,\lambda_2]^k$ of
all return vectors such that for any return vector $\a\in
[\lambda_1,\lambda_2]^k$ an element $\a_i\in \tilde A$ exists
satisfying $\|\a-\a_i\|<\mu$, where $M=(1/\mu)^k$. \footnote{
In what follows we ignore the problem of rounding. }

For any vector $\a\in\tilde A$, let $\delta[\a]=(0,\dots ,~1,\dots ,~0)$
be the probability distribution
concentrated on element $\a$ of the set $\tilde A$.
In this vector of dimension $M$, the $i$th coordinate is 1,
all other coordinates are 0.

Consider a set $P(\tilde A|\tilde C)$ of all probability
distributions on $\tilde A$ conditional with
respect to elements of the set $\tilde C$. Any such distribution
is defined by the $KM$-dimensional vector
$\p=(\p(a_i|c_j): 1\le i\le M, 1\le j\le K)'$, where
$\sum\limits_{i=1}^M\p(a_i|c_j)=1$ for each $1\le j\le K$, i.e.,
given $\c_j\in\tilde C$,
$\p(\cdot |\c_j)=(\p(a_i|c_j: 1\le i\le M)'$ is an $M$-dimensional
probability vector.\footnote{$\x'$ is the transposition of a vector $\x$.
}

Let $\tilde P(\tilde A|\tilde C)=\{\s_1,\dots ,~\s_N)$ be an $\epsilon$-grid in
the set $P(\tilde A|\tilde C)$. For any $\P\in P(\tilde A|\tilde C)$
an $\s_i\in \tilde P(\tilde A|\tilde C)$ exists such that $\|\P-\s_i\|_1<\epsilon$.

Now we apply Theorem~\ref{approach-2}.
Consider a two-players infinitely repeated
game, where the first player moves are elements of the $N$-element set
$I=\tilde P(\tilde A|\tilde C)$ and the second player moves
are from the $KM$-element set
$J=\tilde A\times\tilde C$. At any round $t$ the first player
outputs a forecast $\p_t\in \tilde P(\tilde A|\tilde C)$ and
the second player outputs an outcome $(\x_t,\z_t)\in J$.

Let $\s_i$ be the $i$th vector of the set
$\tilde P(\tilde A|\tilde C)$, $\c_j$ be the $j$th element
of the set $\tilde C$, and $\a\in\tilde A$, $\0^{KM}$ be the $KM$-dimensional
zero vector.
The values of payoff function $\f$ are vectors of dimension $KMN$:
\begin{eqnarray*}
\f(\s_i,(\a,\c_j))=
\left(\begin{array}{cccccc}
\0^{NM}
\\
\dots
\\
\0^{NM}
\\
\g_j(\s_i,(\a,\c_j))
\\
\0^{NM}
\\
\dots
\\
\0^{NM}
\end{array}\right),
\end{eqnarray*}
where $\g_j(\s_i,\a)$ is its $j$th $NM$-dimensional column-vector defined
as a combination of $N$ column-vectors of dimension $M$:
\begin{eqnarray*}
\g_j(\s_i,(\a,\c_j))=
\left(\begin{array}{cccccc}
\0^{M}
\\
\dots
\\
\0^{M}
\\
\delta [\a]-\s_i(\cdot|\c_j)
\\
\0^{M}
\\
\dots
\\
\0^{M}
\end{array}\right),
\end{eqnarray*}
where $\delta [\a]-\s_i(\cdot|\c_j)$ is the difference
of two $M$-dimensional column-vectors, which is the $i$th component of
the composite vector $\g_j(\s_i,(\a,\c_j))$.

We now define the convex set
$U=\{\x\in {\cal R}^{KNM}:\|\x\|_1\le\epsilon\}$ in the space
${\cal R}^{KNM}$.

By definition, a randomized strategy of the first player is a
sequence $\P_1,\P_2,\dots$, where each $\P_t$ is a conditional 
(with respect to past moves of both players) probability
distribution on $I=\tilde P(\tilde A|\tilde C)$.

A set $U$ is approachable if a randomized strategy
$\P_1,\P_2,\dots$ of the first player exists such that
$\lim\limits_{t\to\infty}{\rm dist}(\f(\p_t,(\x_t,\z_t)),U)=0$
almost surely regardless of the second player moves
$(\x_1,\z_1),(\x_2,\z_2),\dots$,
where the trajectory $\p_1,\p_2,\dots$
is distributed according to the overall probability distribution $P$
and $\p_t\in \tilde P(\tilde A|\tilde C)$ for all $t$.

By Theorem~\ref{approach-2} the closed convex set
$U$ is approachable if and only if for each
$\Q\in {\cal P}(\tilde A\times\tilde C)$ an
$\P\in {\cal P}(\tilde P(\tilde A|\tilde C))$ exists such that $\f(\P,\Q)\in U$.

Let $\Q'(\c_j)=\sum_{i=1}^M \Q(\a_i,\c_j)$ be the marginal probability
distribution on $\tilde C$ and $\Q''(\a_i|\c_j) =\Q(\a_i,\c_j)/\Q'(\c_j)$ be
the conditional probability, where $1\le j\le K$.

Let $\s_k\in \tilde P(\tilde A|\tilde C)$ be such that $\|\s_k-\Q''\|_1<\epsilon$
and $\P=\delta[\s_k]$ be the probability
distribution on $\tilde P(\tilde A|\tilde C)$ concentrated on $\s_k$.
Then by definition
$$
\f(\P,\Q)=(\g_1(\P,\Q),\dots,\g_j(\P,\Q),\dots,\g_K(\P,\Q))',
$$
where
\begin{eqnarray*}
\g_j(\P,\Q)=
\left(\begin{array}{cccccc}
\0^M
\\
\dots
\\
\0^M
\\
\Q(\a_1,\c_j)-\s_k(\a_1|\c_j)\sum_{i=1}^M Q(\a_i,\c_j)
\\
\dots
\\
\Q(\a_i,\c_j)-\s_k(\a_i|\c_j)\sum_{i=1}^M Q(\a_i,\c_j)
\\
\dots
\\
\Q(\a_M,\c_j)-\s_k(\a_M|\c_j)\sum_{i=1}^M Q(\a_i,\c_j)
\\
\0^M
\\
\dots
\\
\0^M
\end{array}\right)=
\end{eqnarray*}
%\begin{eqnarray*}
%=\left(\begin{array}{cccccc}
%\0^M
%\\
%\dots
%\\
%\0^M
%\\
%\Q(\a_1,\c_j)-\s_k(\a_1|\c_j)Q'(\c_j)
%\\
%\dots
%\\
%\Q''(\a_i,\c_j)-\s_k(\a_i|\c_j)Q'(\c_j)
%\\
%\dots
%\\
%\Q''(\a_M,\c_j)-\s_k(\a_M|\c_j)Q'(\c_j)
%\\
%\0^M
%\\
%\dots
%\\
%\0^M
%\end{array}\right)=
%\end{eqnarray*}
=
\begin{eqnarray*}
=Q'(\c_j)\left(\begin{array}{cccccc}
\0^M
\\
\dots
\\
\0^M
\\
\Q''(\a_1|\c_j)-\s_k(\a_1|\c_j)
\\
\dots
\\
\Q''(\a_i|\c_j)-\s_k(\a_i|\c_j)
\\
\dots
\\
\Q''(\a_M|\c_j)-\s_k(\a_M|\c_j)
\\
\0^M
\\
\dots
\\
\0^M
\end{array}\right)
\end{eqnarray*}
for $1\le j\le K$. From this
$\|\f(\P,\Q)\|_1=\sum_{j=1}^K\Q'(\c_j)\|\s_k(\cdot|\c_j)-
Q''(\cdot|\c_j)\|_1<\epsilon$ follows and then
$\f(\P,\Q)\in U$.

By a randomized online strategy of {\it Investor} (first
player) we mean a sequence of conditional probability
distributions $\P_t=\P_t(\s|\sigma_t,\z_t)$, $t=1,2,\dots$, on
$\tilde P(\tilde A|\tilde C)$, where $\s\in \tilde P(\tilde
A|\tilde C)$, $\sigma_t=(\z_1,\p_1,\x_1,\dots ,
\z_{t-1},\p_{t-1},\x_{t-1})$ is a history, and $\z_t\in\tilde C$.

By Theorem~\ref{approach-2}
a randomized strategy $\P_1,\P_2,\dots$ of the first player exists,
such that regardless of that sequence $(\x_1,\z_1),(\x_2,\z_2),\dots$
was announced by the
second player the sequence of the vector-valued payoffs
\begin{eqnarray*}
\bar m_t=\frac{1}{T}\sum\limits_{t=1}^T \f(\p_t,(\x_t,\z_t))=
\left(\begin{array}{cccccc}
\frac{1}{T}\sum\limits_{t=1}^T
I_{\{\p_t=\s_1,\z_t=\c_1\}}(\delta [\x_t]-\s_1(\cdot|\c_1))
\\
\dots
\\
\frac{1}{T}\sum\limits_{t=1}^T
I_{\{\p_t=\s_N,\z_t=\c_1\}}(\delta [\x_t]-\s_N(\cdot|\c_1))
\\
\dots
\\
\dots
\\
\frac{1}{T}\sum\limits_{t=1}^T
I_{\{\p_t=\s_1,\z_t=\c_K\}}(\delta [\x_t]-\s_1(\cdot|\c_K))
\\
\dots
\\
\frac{1}{T}\sum\limits_{t=1}^T
I_{\{\p_t=\s_N,\z_t=\c_K\}}(\delta [\x_t]-\s_N(\cdot|\c_K))
\end{array}\right).
\end{eqnarray*}
$P$-almost surely approaches the set $U$, where $P$
is an overall probability distribution generated by a
the sequence $\P_1,\P_2,\dots$ of conditional distributions
and the trajectory $\p_1,\p_2, \dots$ is distributed by the measure $P$;
by $I_{\{\p_t=\s_1,\z_t=\c_j\}}$ we denote the indicator function.

%In the portfolio game, $\x_1,\x_2,\dots$ is a sequence of
%return vectors and $\z_1,\z_2,\dots$ is a sequence of all signals
%announced in the process of the game.

Let $\p_1,\p_2,\dots$ be a sequence of the well-calibrated forecasts
distributed according to $\P_1,\P_2,\dots$. Denote
$$
N_T(\s,i,j)=|\{t:\p_t=\s, 1\le t\le T,\x_t=\a_i,\z_t=\c_j\}|
$$
and
$$
M_T(\s,j)=|\{t:\p_t=\s, 1\le t\le T,\z_t=\c_j\}|,
$$
where $1\le i\le M$, $1\le j\le K$, and
$\s$ be an arbitrary element of the grid $\tilde P(\tilde A|\tilde C)$.

Approachability of the set $U$ implies the following theorem which is an 
immediate corollary of Theorem~\ref{approach-2}.
\begin{theorem}\label{calibra-1}
There exists a randomized online
strategy $\P_1,\P_2,\dots$ such that for any sequence
$\z_1,\x_1,\dots ,\z_t,\x_t,\dots$
\begin{eqnarray}
\limsup_{T\to\infty}\frac{1}{T}\sum\limits_{1\le j\le K}
\sum\limits_{\s\in \tilde P(\tilde A|\tilde C)}
\sum\limits_{1\le i\le M}|N_T(\s,i,j)-M_T(\s,j)s(i|\c_j)|\le\epsilon
\label{well-cal-1}
\end{eqnarray}
for almost all sequences $\p_1,\p_2,\dots$ distributed according to the
overall probability distribution generated by $\P_1,\P_2,\dots$,
where $\s$ is an arbitrary element of $\tilde P(\tilde A|\tilde C)=\{\s_1,\dots ,\s_M\}$.
\end{theorem}
We call forecasts $\p_1,\p_2,\dots$ satisfying (\ref{well-cal-1})
$\epsilon$-calibrated. If (\ref{well-cal-1}) holds for each $\epsilon>0$
then these forecasts are called well-calibrated. 

A difference with
Foster--Vohra~\cite{FoV98} calibration is that each $\p_t$ is a conditional
probability distribution and that this is calibration
with respect to a side information.

Let $\c_j\in\tilde C$ and $\X$ be a random variable distributed according
to the probability
distribution $\s(\cdot|\c_j)=(s(1|\c_j),\dots ,\s(M|\c_j))$, where
$P\{\X=\a_i\}=s(i|\c_j)$ for every $1\le i\le M$.
For any probability distribution
$\s\in \tilde P(\tilde A|\tilde C)=\{\s_1,\dots ,\s_N\}$
and any $\c_j\in\tilde C$ define the optimal portfolio
\begin{eqnarray}
\b^*(\s|\c_j)=\argmax_{{\bf b}}E_{\X\sim\s(\cdot|\c_j)}
(\log(\b\cdot\X)).
\label{univ-1}
\end{eqnarray}
We can rewrite (\ref{univ-1}) also as
$
\b^*(\s|\c_j)=\argmax_{\b}\sum\limits_{i=1}^M
(\log(\b\cdot\a_i))s(i|\c_j).
$

Let $\tilde B_t=\{\b^*(\s_1|\z_t), \dots ,\b^*(\s_N|\z_t)\}$ be
the set of all optimal portfolios, where $\z_t\in\tilde C$ be a
side information at round $t$. Using a randomized forecast
$\p_t\in \tilde P(\tilde A|\tilde C)$ distributed according to
$\P_t$ existing by the Blackwell approachability theorem, we
can define at any round $t$ of the game presented on
Fig~\ref{fig-1} the random portfolio
\begin{eqnarray}
\b^*_t=\b^*(\p_t|\z_t)=\argmax_{\b}E_{\X\sim\p_t(\cdot|\z_t)}(\log(\b\cdot\X)),
\label{univ-1a}
\end{eqnarray}
where $\p_t\in \tilde P(\tilde A|\tilde C)$ is the random forecast announced at round $t$
and $\z_t$ is the signal at step $t$.

The corresponding randomized algorithm is presented on
Fig.~\ref{fig-1}. We consider two types of investors: {\it
Investor} uses the randomized algorithm for computing an
optimal portfolio, {\it Stationary Investor} uses a continuous
function of the side information.

\begin{figure}
\fbox{%
\parbox{12cm}{%
\noindent FOR $t=1,2\dots$
\\
{\it Market} announces a signal $\z_t$.
\\
{\it Investor} announces a probability distribution
$\P_t=\P_t(\cdot|\sigma_t,\z_t)$ on the set $\tilde P(\tilde A|\tilde C)$.
\\
{\it Market} announces a return vector $\x_t$.
\\
{\it Investor} pick up a portfolio $\b_t^*\in \tilde B_t$
distributed by $\P_t$ considered on $\tilde B_t$
and updates his wealth $S^*_t=S^*_{t-1}\cdot(\b^*_t\cdot\x_t)$.
\\
{\it Stationary Investor} updates his wealth
$S_t=S_{t-1}\cdot(\b(\z_t)\cdot\x_t)$.
\\
ENDFOR
}%
}
\caption{Protocol of portfolio game}\label{fig-1}
\end{figure}

We have presented the construction using the approximation
grids of a fixed accuracy. The complete construction is divided
on time intervals $1\le t_1\le\dots <t_n\le\dots$. At time
steps $t\in [t_n,t_{n+1})$ the grids $\tilde C_n$, $\tilde A_n$
of cardinality $K_n$, $M_n$ and $\tilde P_{n}(\tilde
A_{n}|\tilde C_{n})$ of cardinality $N_n$ are used, where
$n=1,2,\dots$. These grids approximate the sets $C$, $A$, and
$P(A)$ with an increasing degree of accuracy: $M_n\to\infty$,
$K_n\to\infty$ and  $\epsilon_n\to 0$, $\mu_n\to 0$ as
$n\to\infty$.

The following theorem asserts that in this case portfolio (\ref{univ-1a}) is almost surely
log-optimal with respect to the class of all portfolios presented by
continuous functions $\b(\cdot):C\to\cal R$.
\begin{theorem}\label{opp-1}
The randomized portfolio strategy $\b^*_t$ defined by
(\ref{univ-1a}) and by refining the discretization
incrementally is almost surely log-optimal for the class of all
continuous portfolio strategies $\b(\cdot)$:
\begin{eqnarray}
\liminf\limits_{T\to\infty}\frac{1}{T}\log\frac{S_T^*}{S_T}\ge 0
 \label{opt-1}
\end{eqnarray}
for almost all trajectories $\p_1,\p_2,\dots$, where
$S^*_T=\prod_{t=1}^T(\b^*_t\cdot\x_t)$ is the wealth
achieved by of the universal portfolio strategy,
$S_T=\prod_{t=1}^T(\b(\z_t)\cdot\x_t)$ is
the wealth achieved by an arbitrary portfolio $\b(\z_t)$, and
$\z_t$ is a signal at any round $t$.
\end{theorem}
{\it Proof}. The complete proof is based on the construction which is divided on
time intervals $1\le t_1\le\dots <t_n\le\dots$ and the grids
with increasing degree of accuracy as indicated above.
For simplicity of presentation, we give the proof only for one of such grids.
Given $\epsilon>0$ and $\mu>0$, we consider the
corresponding grids $\tilde C$, $\tilde A$, and $\tilde P(\tilde A|\tilde C)$
and prove optimality of the universal portfolio up to $O(\mu+\epsilon)$.
Notice that $M=(1/\mu)^k$ and $N=(1/\epsilon)^{KM-1}$.

Replace the sums with return vectors and signals
on their approximations from the corresponding grids.
Let us estimate the loss of accuracy as a result of such replacements.

Notice that for any $\b\in\Gamma$, $\x\in A$ and $\a\in\tilde A$
such that $\|\x-\a\|<\mu$ we have
%\begin{eqnarray*}
$
|(\b\cdot\x)-(\b\cdot\a)|=|(\b\cdot (\x-\a))|\le \|\b\|\|\x-\a\|\le
\|\x-\a\|<\mu.
$
%\end{eqnarray*}
Then for $\b\cdot\x)\ge (\b\cdot\a)$,
%\begin{eqnarray*}
$
|\ln(\b\cdot\x)-\ln(\b\cdot\a)|=\left|\ln\frac{(\b\cdot\x)}{(\b\cdot\a)}\right|=
%\\
\left|\ln\left(1+\frac{(\b\cdot\x)-(\b\cdot\a)}{(\b\cdot\a)}\right)\right|\le
\left|\frac{(\b\cdot\x)-(\b\cdot\a)}{(\b\cdot\a)}\right|\le\mu/\lambda_1,
$
%\end{eqnarray*}
where $\ln$ is the natural logarithm; if $\b\cdot\x)<(\b\cdot\a)$
we can exchange role of $\a$ and $\x$.

Let $\b(\cdot)$ be an arbitrary continuous stationary portfolio strategy.
Given $\epsilon>0$ consider a sufficiently accurate approximating grid
$\tilde C=\{\c_1,\dots ,\c_K\}$ in
the set $C$ of all signals satisfying the following property:
for each $\z\in C$ an $\c_i\in\tilde C$ exists such that\footnote
{
Since the complete construction is based on the sequence of $\epsilon_k$-grids,
where $\epsilon_k\to 0$ as $k\to\infty$, for each continuous function
$\b(\cdot)$ an $\epsilon_k$-grid exists such that this property holds.
}
$\|\b(\z)-\b(\c_i)\|<\epsilon$.

Therefore, all sums are changed on $O(\mu+\epsilon)$ as a result of these
replacements of return vectors and signals. Assuming that return vectors
$\x_t$ and signals $\z_t$ are now elements of the corresponding
finite grids and using continuity of the function $\b(\cdot)$, we obtain
the estimate for the mean wealth of an arbitrary portfolio $\b(\cdot)$:
\begin{eqnarray}
\frac{1}{T}\log S_T=
\frac{1}{T}\sum\limits_{t=1}^T\log (\b(\z_t)\cdot\x_t)=
\nonumber
\\
\sum\limits_{j=1}^K
\frac{1}{T}\sum\limits_{\s\in \tilde P(\tilde A|\tilde C)}
\sum\limits_{i=1}^M
N_T(\s,i,j)\log (\b(\c_j)\cdot\a_i)+O(\epsilon+\mu),
\label{tru-2}
\end{eqnarray}
where $N_T(\s,i,j)=|\{t:\p_t=\s, 0\le t\le T,\x_t=\a_i, \z_t=\c_j\}|$.
By (\ref{well-cal-1}) of Theorem~\ref{calibra-1}
$$
\frac{1}{T}\sum\limits_{1\le j\le K}\sum\limits_{\s\in \tilde P(\tilde A|\tilde C)}
\sum\limits_{1\le i\le M}|N_T(\s,i,j)-M_T(\s,j)s(i|\c_j)|\le\epsilon+o(1)
$$
as $T\to\infty$ almost surely.

Starting from (\ref{tru-2}) we obtain the following chain of equalities
and inequalities which are valid almost surely
\begin{eqnarray}
\frac{1}{T}\sum\limits_{1\le j\le K}\sum\limits_{\s\in \tilde P(\tilde A|\tilde C)}
\sum\limits_{i=1}^M
N_T(\s,i,j)\log (\b(\c_j)\cdot\a_i)=
\nonumber
\\
\frac{1}{T}\sum\limits_{1\le j\le K}\sum\limits_{\s\in \tilde P(\tilde A|\tilde C)}
M_T(\s,j)\sum\limits_{i=1}^M
\log (\b(\c_j)\cdot\a_i)s(i|\c_j)+O(\epsilon)+o(1)=
\nonumber
\\
\frac{1}{T}\sum\limits_{1\le j\le K}\sum\limits_{\s\in \tilde P(\tilde A|\tilde C)} M_T(\s,j)
E_{\X\sim\s(\cdot|\c_j)}(\log (\b(\c_j)\cdot\X))+O(\epsilon)+o(1)\le
\nonumber
\\
\frac{1}{T}\sum\limits_{1\le j\le K}\sum\limits_{\s\in \tilde P(\tilde A|\tilde C)}
M_T(\s,j)E_{\X\sim(\s(\cdot|\c_j)}
(\log (\b^*(\s|\c_j)\cdot\X))+O(\epsilon)+o(1)=
\nonumber
\\
\frac{1}{T}\sum\limits_{1\le j\le K}\sum\limits_{\s\in \tilde P(\tilde A|\tilde C)}
M_T(\s,j)\sum\limits_{i=1}^M
\log (\b^*(\s|\c_j)\cdot\a_i)s(i|\c_j)+O(\epsilon)+o(1)=
\nonumber
\\
\frac{1}{T}\sum\limits_{1\le j\le K}\sum\limits_{\s\in \tilde P(\tilde A|\tilde C)}
\frac{1}{T}\sum\limits_{i=1}^M
N_T(\s,i,j)\log (\b^*(\s|\c_j)\cdot\a_i)+O(\epsilon)+o(1).
\label{errr-1}
\end{eqnarray}
as $T\to\infty$.

Now, we change from (\ref{errr-1}) to general setting
\begin{eqnarray}
\sum\limits_{j=1}^K \sum\limits_{\s\in \tilde P(\tilde A|\tilde C)}
\frac{1}{T}\sum\limits_{i=1}^M
N_T(\s,i,j)\log (\b^*(\s|\c_j)\cdot\a_i)+O(\mu+\epsilon)+o(1)=
\nonumber
\\
\frac{1}{T}\sum\limits_{t=1}^T
\log (\b^*_t\cdot\x_t)+O(\mu+\epsilon)+o(1)=
\frac{1}{T}\log S^*_T+O(\mu+\epsilon)+o(1)
\label{errr-2}
\end{eqnarray}
almost surely, where $S^*_T$ is the wealth achieved by the optimal portfolio.

Relations (\ref{tru-2}) and (\ref{errr-2}) imply that almost
surely
$\liminf\limits_{T\to\infty}\frac{1}{T}\log\frac{S^*_T}{S_T}
\ge -c(\epsilon+\mu)$, where $c$ is a positive constant. The
proof for the case of a fixed approximating grid is complete.

\section{Some remarks on rate of convergence}\label{rate-1}

In this section following Mannor and Stoltz~\cite{MaS2009} we discuss rate of
convergence in the optimality condition (\ref{opt-1}). This
rate is defined by the rate of convergence of a calibrated
forecaster in the Blackwell approachability theorem
(\ref{well-cal-1}) and on the infinite series of grids $\tilde
C$, $\tilde A$, and $\tilde P(\tilde A|\tilde C)$. We assume that
all portfolios functions $\b(\cdot)$ are Lipschitz continuous.

The proof of the approachability theorem gives rise to an implicit strategy,
as indicated in Blackwell~\cite{Ble56}. We start from a variant the Blackwell
theorem for $l_2$ norm $\|\cdot\|_2$. Denote
$d_U(\x)$ the projection of $\x$ in $l_2$-–norm onto $U$.
According to the proof of this theorem (see Blackwell~\cite{Ble56} or
Cesa-Bianchi and Lugosi~\cite{LCB2007}, Section 7),
at each round $t>2$ and with the notations above, the forecaster should
pick his action $\p_t$ at random according to a distribution $\P_t$
on the set $\tilde P(\tilde A|\tilde C)$ such that
$
(\bar m_{t-1}-d_U(\bar m_{t-1}))\cdot (\f(\P_t,(\a_j,\c_k))-\bar m_{t-1})\le 0
$
for all $1\le j\le M$ and $1\le k\le K$.

Proof of the Blackwell theorem from Blackwell~\cite{Ble56} and convergence
theorem for Hilbert space-valued martingales
of Kallenberg and Sztencel~\cite{KaS91} (see also Chen and White~\cite{ChW96})
provide uniform convergence rates
of sequence of empirical payoff vectors $\bar m_t$
to the target set $U$: there exists an absolute constant $c$
such that for any $\delta>0$ for all strategies of {\it Market} and for all $T$,
with probability $1-\delta$,
\begin{eqnarray}
\|\bar m_t-d_U(\bar m_t)\|_2\le
c\sqrt{\frac{\log\frac{1}{\delta}}{T}},
\label{delt-11}
\end{eqnarray}
where $\|\cdot\|_2$ is the Euclidian norm in ${R}^{KNM}$.\footnote{
See Cesa-Bianchi and Lugosi~\cite{LCB2007}, Exersice 7.23. Recall that
$N=|\tilde P(\tilde A|\tilde C)|$, $M=|\tilde A|$, and $K=|\tilde C|$.}

The set $U$ used in the proof of Theorem~\ref{opp-1}
is defined using $l_1$-norm. Then, using triangle inequality
and the Cauchy-Schwarz inequality, we obtain
\begin{eqnarray*}
\|\bar m_T\|_1\le\|d_U(\bar m_T\|_1+\|\bar m_T-d_U(\bar m_T)\|_1\le
\epsilon+\sqrt{KMN}\|\bar m_T-d_U(\bar m_T)\|_2,
%\label{delt-1111}
\end{eqnarray*}
where $N=O((1/\epsilon)^{KM-1})$. Hence, by (\ref{delt-11})
we have
\begin{eqnarray}
\|\bar m_T\|_1\le c\epsilon+c
\sqrt{\frac{\log\frac{1}{\delta}}{\epsilon^{KM-1}T}}.
\label{delt-1111}
\end{eqnarray}
The suitable choice of $\epsilon$ is when both terms of the sum (\ref{delt-1111})
are of the same order of magnitude, i.e.,
$\epsilon\sim T^{-\frac{1}{KM+1}}$. This is the optimal level of precision in
(\ref{errr-1}).

Further, we have to find the optimal level of precision in
(\ref{tru-2}). We should optimize the choice of $M=(1/\mu)^k$,
where $\mu$ is the level of precision of of approximating the
values $\x_t$.

Assuming that the signal space $C$ is a closed interval of real
numbers, $\b(\cdot)$ is Lipschitz continuous, and getting
$K=1/\mu$, we should choose the suitable $\mu$ to minimize the
sum $\mu+T^{-\frac{1}{KM+1}}$, where $KM=(1/\mu)^{k+1}$. The
optimal value is $\mu\sim (\log T)^{-\frac{1}{k+2}}$.

Combining series of grids,
like it was done in in Vyugin~\cite{Vyu2013},
we can obtain a rough bound $O(\log T)^{-\frac{1}{k+2}+\nu}$
of the rate of convergence in (\ref{opt-1}),
where $\nu$ is an arbitrary small positive real number, and $k$ is the number
of assets.  More precise, for
any $\delta>0$, with probability $1-\delta$,
$$
\frac{1}{T}\log S^*_T\ge
\frac{1}{T}\log S_T-c(\log (T/\log\frac{1}{\delta}))^{-\frac{1}{k+2}+\nu}
$$
for all $T$, where $c$ is a constant and
$S_T=\prod_{t=1}^T(\b(\z_t)\cdot\x_t)$ is the wealth
achieved by an arbitrary Lipschitz continuous portfolio
$\b(\cdot)$.

%------------------------------------------------------

\section{Competing with discontinuous stationary strategies}\label{discr-1}

The portfolio strategy $\b^*_t$ defined in Theorem~\ref{opp-1}
performs asymptotically at least as well as any continuous strategy $\b(\z_t)$.
A weak point the strategy $\b(\z_t)$ is that a continuous function
cannot respond sufficiently quickly to information about changes of
the return vectors.

A positive argument in favor of the requirement of continuity of $\b(\cdot)$ is that
it is natural to compete only with computable trading strategies,
and continuity is often regarded as a necessary condition for computability
(Brouwer's ``continuity principle'').

If $\b(\cdot)$ is allowed to be discontinuous, we cannot prove
asymptotic optimality of our portfolio strategy $\b_t^*$.
We present below the corresponding construction.

Consider a portfolio game with two assets and suppose that an
algorithm exists which when fed with $t$ and a history
$\sigma_t=\b_1^*,\x_1,\dots ,\b_{t-1}^*,\x_{t-1}$ outputs a
probability distribution $P_t=P_t(\cdot|\sigma_t)$ on the
simplex of all portfolios and let a portfolio $\b^*_t$ be
distributed by $P_t$. Denote by
$\e_t=E_{P_t}(\b^*_t)=(e_{1,t},e_{2,t})'$ the conditional
mathematical expectation of $\b^*_t$ with respect to
$\sigma_t$.

Suppose that $\z_t=\sigma_t$ be a signal at round $t$.
Define $\b(\z_t)=(b_1(\z_t),b_2(\z_t))'$, where
\[
b_1(\z_t)=
\left\{
    \begin{array}{l}
      1 \mbox{ if } e_{1,t}\le\frac{1}{2}
    \\
      0 \mbox{ otherwise}
    \end{array}
  \right.
\]
and $b_2(\z_t)=1-b_1(\z_t)$.

For any $t$, define a return vector $\x_t=(x_{1,t},x_{2,t})'$,
where $x_{1,t}=2$, $x_{2,t}=1$ if $e_{1,t}\le\frac{1}{2}$ and
$x_{1,t}=1$, $x_{2,t}=2$ otherwise.

Let $S_T=\prod_{t=1}^T(\b(\z_t)\cdot\x_t)$ be a wealth achieved
by the stationary portfolio $\b(\cdot)$ and
$S_T^*=\prod_{t=1}^T(\b_t^*\cdot\x_t)$ be a wealth achieved
by the randomized portfolio strategy $\b_t^*$ in the first $T$ rounds.

By definition $\log\frac{S_T}{S^*_T}=
\sum\limits_{t=1}^T\log\frac{(\b(\z_t)\cdot\x_t)}{(\b^*_t\cdot\x_t)}$.
It is easy to verify that for any $t$,
\[
\log\frac{(\b(\z_t)\cdot\x_t)}{(\b^*_t\cdot\x_t)}=
\left\{
    \begin{array}{l}
      \frac{2}{1+b^*_{1,t}} \mbox{ if } e_{1,t}\le\frac{1}{2}
    \\
      \frac{2}{1+b^*_{2,t}} \mbox{ otherwise }
    \end{array}
  \right.
\]
Then for any $t$,
\[
E_{P_t}\left(\log\frac{(\b(\z_t)\cdot\x_t)}
{(\b^*_t\cdot\x_t)}\right)=
\left\{
    \begin{array}{l}
      E_{P_t}\frac{2}{1+b^*_{1,t}}\ge\frac{1}{2}
E_{P_t}\frac{1-b^*_{1,t}}{1+b^*_{1,t}}\ge\frac{1}{8}
 \mbox{ if } e_{1,t}\le\frac{1}{2}
    \\
      E_{P_t}\frac{2}{1+b^*_{2,t}}\ge
\frac{1}{2}
E_{P_t}\frac{1-b^*_{2,t}}{1+b^*_{2,t}}\ge\frac{1}{8}
\mbox{ otherwise }
    \end{array}
  \right.
\]
Then $E_{P}\left(\log\frac{S_T}{S^*_T}\right)\ge\frac{1}{8}T$
for all $T$, where $P$ is an overall probability distribution
defined by all $P_t$, $t=1,2,\dots$. By the martingale law of
large numbers
$$
\liminf\limits_{T\to\infty}\frac{1}{T}\log\frac{S_T}{S^*_T}\ge\frac{1}{8}
$$
almost surely. Therefore, the stationary portfolio strategy $\b(\cdot)$
outperforms the portfolio strategy $\b_t^*$ almost surely.
%----------------------------------------------------------

\section{Conclusion}

In this paper we show how the game-theoretic methods can be
applied to the classical problems of a universal portfolio
construction. We present the method for constructing a
log-optimal portfolio in a game-theoretic framework and in
adversarial setting. No stochastic assumptions are made about
return vectors. Instead, we define ``an artificial probability
distribution'' for return vectrs using the method of
calibration. Using this distribution, we construct the
log-optimal portfolio by the standard scheme
(\ref{log-optimal-1}), where the mathematical expectation $E$
is over probability distribution defined by well-calibrated
forecasts. Our log-optimal portfolio performs asymptotically at
least as well as any stationary portfolio that redistribute the
investment at each round using a continuous function of the
side information. This performance is almost surely, where the
corresponding probability distribution is an internal
distribution of the probabilistic algorithm computing
well-calibrated forecasts on the base of the Blackwell
approachability theorem. Theorem~\ref{opp-1} is valid not only
for log-loss function but also for any Lipschitz continuous
loss-function.

The drawback of this approach is the very poor rate of
convergence in~(\ref{opt-1}), but it is true for basically all
nontrivial applications of approachability 
(probably except of V'yugin~\cite{Vyu2013}) because the
constructions used to employ it results in many extra
dimensions which do not suit $l_2$ geometry for the problem.
Note that no rate of convergence exists for portfolio
strategies universal with respect to the class of all
stationary and ergodic processes.

\section{Acknowledgement}
This research was partially supported by Russian foundation
for fundamental research: Grant 13-01-12447 and 13-01-00521.

\end{document}